\documentclass[12pt]{article}
\usepackage{hyperref}
\usepackage{url}
\usepackage[utf8]{inputenc} 
\usepackage[T1]{fontenc}
\usepackage[italian]{babel}
\usepackage{graphicx}
\usepackage{multibib}

\newcites{altre}{Altre pubblicazioni utili}
\usepackage{enumerate}
\usepackage{lmodern}
\begin{document}
\setcounter{tocdepth}{1}
\title{Introduzione all’Intelligenza Artificiale\footnote{Una versione precedente di questo articolo è stato pubblicato sulla rivista Terre di Confine, Anno 2, Numero 1, gennaio 2006, \url{http://www.terrediconfine.eu/introduzione-all-intelligenza-artificiale/}}}

\author{Fabrizio Riguzzi\\
Dipartimento di Matematica e Informatica -- Università di Ferrara\\ Via Saragat 1, I-44122, Ferrara, Italy\\\texttt{fabrizio.riguzzi@unife.it}}

\date{gennaio 2006} 
\maketitle
\nocitealtre{car}
\nocitealtre{con}
\nocitealtre{gor}
\nocitealtre{odi}

\abstract{L'articolo presenta un'introduzione all'Intelligenza Artificiale (IA) in forma
divulgativa e informale ma precisa. L'articolo affronta prevalentemente gli aspetti
informatici della disciplina, presentando varie tecniche usate nei sistemi di IA e 
dividendole in simboliche e subsimboliche. L'articolo si conclude presentando 
il dibattito in corso sull'IA e in particolare sui vantaggi e i pericoli che 
sono stati individuati, terminando con l'opinione dell'autore al riguardo.}
 
\section{Introduzione}
         Il termine Intelligenza Artificiale (d’ora in avanti IA) è stato inventato da John McCarthy nel 1956, in occasione di un seminario di due mesi (da lui organizzato al Dartmouth College di Hanover, New Hampshire, USA) che ebbe il merito di far conoscere tra loro 10 studiosi (su teoria degli automi, reti neurali e intelligenza) statunitensi, e di dare l’imprimatur al termine “Intelligenza Artificiale” come nome ufficiale del nuovo campo di ricerca.
				
      Da allora l’IA si è affermata ed evoluta; oggi è riconosciuta come branca autonoma, sebbene connessa a informatica, matematica, scienze cognitive, neurobiologia e filosofia.
      
			Molte definizioni sono state date di questa materia: esse differiscono per i compiti svolti dalle macchine che l’IA cerca di costruire. Tali compiti possono essere classificati secondo due dimensioni ortogonali \cite{rusnor}: macchine che pensano o agiscono, e macchine che simulano gli umani (o che si comportano razionalmente). In tutto quattro classi, a seconda che le macchine: pensino come umani, agiscano come umani, pensino razionalmente, agiscano razionalmente. Quattro approcci distinti alla ricerca nel campo dell’IA, dunque, tutti attivamente perseguiti.
      
			L’obiettivo dell’approccio “macchine che pensano come umani” è quello di riprodurre il ragionamento umano nelle macchine. Può essere fatto a due livelli: imitando i metodi di ragionamento o replicando il funzionamento del cervello. Nel primo caso, la scienza cognitiva ci fornisce un importante punto di partenza, ottenuto mediante introspezione ed esperimenti psicologici. Nel secondo caso, è la neurobiologia a fornirci un modello adeguato. Questo primo criterio si occupa quindi di produrre macchine automi che, oltre a comportarsi come umani, “funzionino” come umani tali.
      
			L’obiettivo dell’approccio “macchine che si comportano come umani” è quello di realizzare macchine indistinguibili dagli uomini. Questa proprietà è stata meglio definita da Alan Turing che, in un suo articolo del 1950 \cite{tur}, ha proposto il test che prende il suo nome: un “giudice” ha la facoltà di porre a un “soggetto” domande per iscritto e, in base alle risposte, deve decidere se si tratta di un uomo o di una macchina. Al fine di superare il test di Turing, una macchina deve esibire le seguenti capacità:
			\begin{itemize}
			\item
elaborazione del linguaggio naturale, al fine di comunicare efficacemente nella lingua del giudice;
\item rappresentazione della conoscenza, per memorizzare quello che sa o impara;
\item ragionamento automatico, per inferire (produrre), a partire dalla propria conoscenza, le risposte al giudice;
\item apprendimento automatico, per aumentare la propria base di conoscenza.
\end{itemize}
      Il test di Turing non prevede interazione fisica tra il giudice e la macchina, non essendo necessaria. Volendo, si può pensare a un test di Turing totale in cui, invece delle risposte scritte, il giudice riceve un segnale audio-video, ed ha la possibilità di passare degli oggetti alla macchina attraverso una feritoia. In questo caso la macchina deve esibire anche le seguenti capacità:
      \begin{itemize}
		\item	visione artificiale, per riconoscere gli oggetti ricevuti;
		\item	robotica, per manipolarli;
\item	elaborazione del linguaggio parlato, per comprendere le domande del giudice e per rispondere.
\end{itemize}
      Il test di Turing totale ricorda il test Voight-Kampf nel film Blade Runner con il quale i poliziotti distinguono gli androidi dagli esseri umani.
      
			L’approccio “macchine che pensano razionalmente” non si preoccupa che le macchine realizzate funzionino come umani, ma solo che seguano ragionamenti razionali, dove “razionale” è definito in maniera precisa dalla matematica, anche tramite tecniche che gli esseri umani naturalmente non usano. Ad esempio la logica, cioè lo studio di come effettuare ragionamenti inattaccabili. La logica svolge un importante ruolo nell’IA, anche se le aspettative iniziali sono state ridimensionate dai limiti pratici del suo uso, soprattutto in situazioni di conoscenza incompleta e/o incerta.
      
			L’approccio “macchine che agiscono razionalmente” usa la definizione di “azione razionale” fornita dall’economia, ossia: selezione delle azioni che portano al migliore risultato, o al migliore risultato atteso nel caso ci siano elementi di impredicibilità. L’obiettivo di questo approccio è quello di realizzare un agente, un'entità in grado di agire in un ambiente al fine di raggiungere uno o più obbiettivi. L’agente utilizzerà il ragionamento razionale per scegliere quali azioni compiere, ma in alcuni casi dovrà reagire agli stimoli ambientali in maniera tanto veloce da “scavalcare” la scelta (ad esempio quando una inazione mettesse a rischio la sua esistenza). Se si tocca qualcosa che scotta, si reagisce ritirando immediatamente la mano, senza un ragionamento cosciente; allo stesso modo l’agente, in certe situazioni, deve poter agire senza svolgere un ragionamento. Gli agenti possono essere di due tipi: solo software, e in questo caso si chiamano softbot, o sia hardware che software, chiamati allora robot. Nel caso dei softbot, l’ambiente esterno in cui operano è rappresentato da Internet, dove interagiscono con esseri umani e altri softbot. Questo, al momento, è l’approccio maggiormente perseguito, in quanto quello che promette i risultati di maggior utilità pratica. 
      
			Al fine di presentare le varie tecniche proposte dall’IA, le suddivideremo in due grandi classi: simboliche e subsimboliche. Le prime si propongono di automatizzare il ragionamento e l’azione, rappresentando le situazioni oggetto di analisi tramite simboli comprensibili agli esseri umani, ed elaborandole mediante algoritmi. Le seconde, invece, non rappresentano esplicitamente la conoscenza in maniera direttamente comprensibile, e sono basate sulla riproduzione di fenomeni naturali.
			
			Le tecniche simboliche sono state il paradigma dominante nell'IA da metà degli anni 50 alla fine degli anni 80 e sono state chiamate 
			 GOFAI (“Good Old-Fashioned Artificial Intelligence”) in \cite{10.5555/4694}. Le tecniche subsimboliche, e in particolare le reti neurali, hanno preso piede a partire dagli anni 90 e hanno ottenuto 
			 importanti successi in vari campi come ad esempio la visione artificiale. Recentemente la ricerca si sta orientando verso la combinazione di tecniche simboliche e subsimboliche.
      
\section{Tecniche simboliche}
      Tra le tecniche simboliche descriveremo la ricerca nello spazio degli stati, il ragionamento automatico e l’apprendimento automatico.

\subsection{Ricerca nello spazio degli stati}
      Questa tecnica viene utilizzata quando si vuole scegliere una serie di azioni che portino da uno stato iniziale a uno o più stati finali desiderati. Condizioni, affinché possa essere utilizzata, sono: che lo stato del mondo esterno possa essere rappresentato in maniera concisa (in forma simbolica), che le azioni disponibili possano essere espresse come regole per il passaggio da uno stato al successivo e che esista un test per stabilire se uno stato è finale. 
			
\begin{figure}[t]
\centering
\includegraphics[width=.45\textwidth]{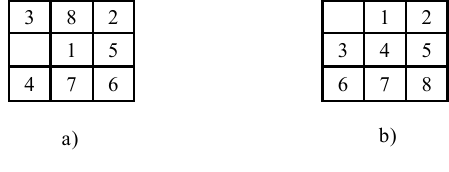}
\caption{Stati iniziale e finale del gioco dell'otto.\label{otto}}
	\end{figure}

      Vediamo un esempio di problema che può essere trattato in questo modo. Nel gioco dell’otto (o puzzle dell’otto) si ha una scacchiera tre per tre, in cui otto caselle sono occupate da otto tessere numerate dall’1 all’8 e una casella è vuota. Si parte da una configurazione iniziale casuale delle tessere, come ad esempio quella rappresentata in Figura \ref{otto} a), e si vuole arrivare alla configurazione rappresentata in Figura \ref{otto} b).

			Le mosse possibili consistono nello spostare sulla casella vuota una tessera numerata ad essa adiacente. Per questo problema, lo “stato” consiste nella posizione delle otto tessere numerate: lo stato iniziale è quello rappresentato in Figura \ref{otto} a), e il test per stabilire se si è raggiunto uno stato finale verifica semplicemente che lo stato sia identico a quello di Figura \ref{otto} b).
    
\begin{figure}[t]
\centering
\includegraphics[width=.9\textwidth]{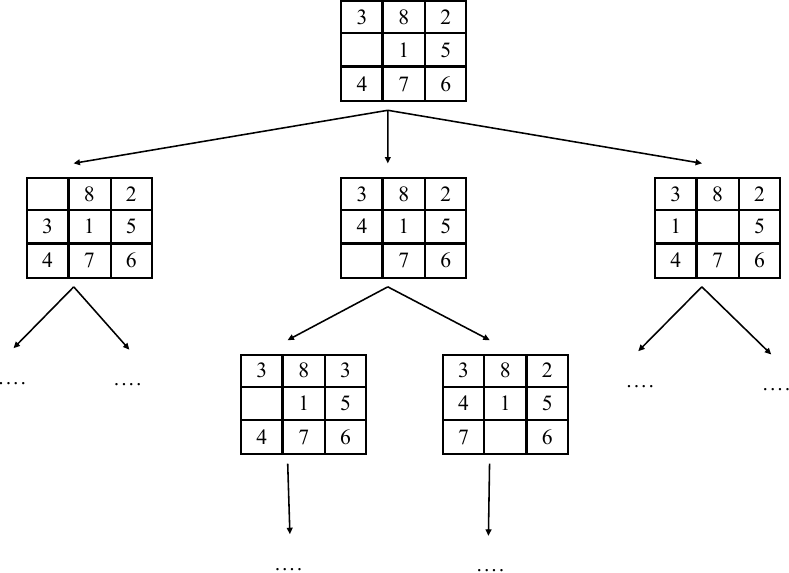}
\caption{Spazio degli stati del gioco dell'otto.\label{ssotto}}
	\end{figure}

			Si tratta di un problema che sembra richiedere intelligenza: un essere umano lo risolverebbe provando diverse mosse e cercando di prevederne il risultato. Il metodo di soluzione proposto dall’IA consiste nel compiere una ricerca nello spazio dei possibili stati. A tal fine, si può rappresentare lo “spazio” come un albero in cui ogni nodo corrisponde a uno “stato”. La radice dell’albero è lo stato iniziale, i figli di un nodo sono quegli stati che si possono raggiungere dallo stato associato al nodo applicando una sola mossa. Ad esempio, nel caso del gioco dell’otto, lo spazio degli stati si può rappresentare come l’albero di Figura \ref{ssotto}. 
      
			Il problema è risolto quando si è trovato un percorso dallo stato iniziale a uno stato finale. Solitamente non basta trovare una soluzione ma si cerca quella che ha costo minimo. Occorre quindi definire un “costo”: di norma si assegna un costo alle varie mosse, e il costo di un cammino si misura come somma dei costi delle mosse che lo compongono. Nel caso del gioco dell’otto, ogni mossa costa 1, e si cerca la soluzione che richiede il numero minimo di mosse.
      
			Citiamo altri problemi che si possono risolvere con una ricerca nello spazio degli stati. Iniziamo da quelli “giocattolo”, ovvero problemi che non hanno immediato interesse pratico. 
      
			Il problema “dei missionari e dei cannibali” consiste nel far attraversare un fiume a 3 missionari e 3 cannibali, utilizzando una barca ed evitando che i primi mangino i secondi (quando li superano in numero su una delle due sponde). La barca può contenere al massimo due persone alla volta e, per muoversi da una sponda all’altra del fiume, dev’essere guidata da almeno una persona. Lo stato di questo problema può essere rappresentato utilizzando una terna di numeri, nella quale i primi due quantifichino i missionari e i cannibali sulla sponda iniziale, e il terzo equivalga a: 1 se la barca è sulla sponda iniziale, e 0 se è sull’altra. Lo stato iniziale è (3,3,1), lo stato finale è (0,0,0) e un possibile stato (2,2,1) indica che sulla sponda iniziale ci sono due missionari e due cannibali e che la barca è sulla sponda iniziale. Le operazioni possibili sono cinque: attraversa il fiume con 2 missionari, con 2 cannibali, con un missionario e un cannibale, con un solo cannibale o con un solo missionario. Non tutte le operazioni sono consentite in ogni stato: ad esempio, a partire dallo stato (2,2,1) non è possibile applicare l’operazione “attraversa il fiume con un missionario” perchè sulla sponda iniziale rimarrebbero un missionario e due cannibali, e quindi i cannibali mangerebbero il missionario. Il costo, in questo caso, è unitario per tutte le operazioni, si cercano quindi soluzioni con il minimo numero di attraversamenti del fiume.
      
			Il problema “delle $n$-regine” consiste nel disporre, su una scacchiera $n$ per $n$, $n$ regine in modo che non si attacchino. Una regina attacca tutti i pezzi che si trovano sulla stessa colonna, riga o diagonale. Lo stato in questo caso è rappresentato dalla posizione di $i$ regine sulla scacchiera, con $i$ che va da 0 a $n$. Lo stato iniziale è rappresentato dalla scacchiera senza regine, lo stato finale da una scacchiera con $n$ regine che non si attaccano a vicenda. Le mosse sono l’aggiunta di una regina su una casella della scacchiera. In questo caso tutte le soluzioni hanno uguale costo perchè richiedono tutte 8 mosse.
      
			Passiamo ai problemi reali, come quello di “trovare un percorso”, che consiste nell’andare da un luogo ad un altro con il minimo costo, transitando per luoghi intermedi, al collegamento tra ognuno dei quali è associato un costo. Un esempio è il tragitto in macchina da una città ad un’altra: i collegamenti sono le strade e il costo può essere la distanza o il tempo necessario a percorrere quel collegamento. Se lo spostamento avviene in aereo, i collegamenti sono i voli disponibili, e il costo può essere il tempo o il prezzo del volo. Gli “stati” sono i luoghi, e le mosse disponibili consistono nell’utilizzare uno dei collegamenti che partono dal luogo corrente per spostarsi in un altro luogo. 
      
			Altro problema reale risolvibile mediante ricerca nello spazio degli stati è quello del “tour”, cioè un percorso che parta da un luogo e vi ritorni dopo aver “visitato” almeno una volta tutti i luoghi di un insieme. Questo problema ha le stesse mosse del precedente ma stati diversi: qui lo stato è costituito dal luogo corrente più l’insieme dei luoghi già visitati. Lo stato finale è quindi quello in cui il luogo corrente è il luogo iniziale e l’insieme di luoghi visitati corrisponde a quello specificato. Il costo può essere, anche in questo caso, rappresentato dalla distanza, dal tempo o dal costo monetario delle mosse.
      
			Come si risolvono problemi di ricerca? Occorre generare e percorrere lo spazio di ricerca dal nodo iniziale fino a trovare uno stato che superi mediante verifica il test di stato finale; quando la verifica non viene superata, occorre “espandere” il nodo, generare i suoi successori utilizzando le possibili mosse od operatori, ed esaminare quelli, fino a trovarne uno che superi il test, o fino a che non si sia esplorato tutto lo spazio degli stati. Si genera così progressivamente l’albero di ricerca.
      
			Gli algoritmi per la ricerca nello spazio degli stati differiscono nella scelta del nodo da espandere (strategia di ricerca). Le due strategie più comuni sono la ricerca in profondità e la ricerca in ampiezza.

\begin{figure}[t]
\centering
\includegraphics[width=.95\textwidth]{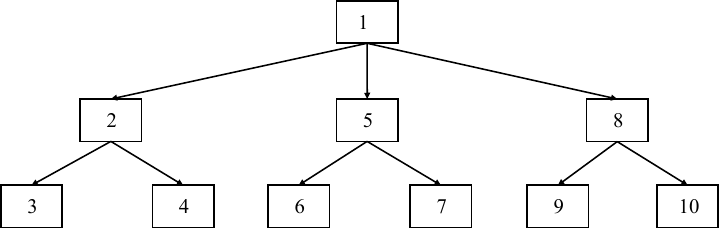}
\caption{Strategia di ricerca in profondità.\label{prof}}
	\end{figure}

\begin{figure}[t]
\centering
\includegraphics[width=.95\textwidth]{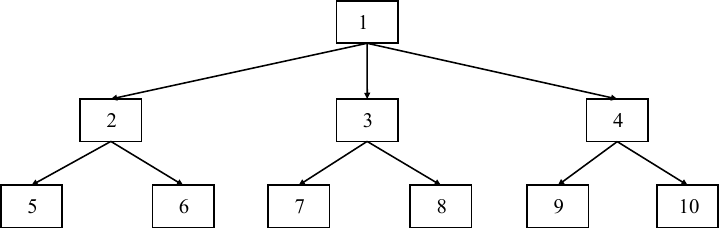}
\caption{Strategia di ricerca in ampiezza.\label{amp}}
	\end{figure}

			Nella ricerca in profondità si espande sempre il nodo a profondità maggiore che sia stato generato ma non ancora espanso. Questo significa che si procede prima in profondità fino ad arrivare ad un nodo che non può essere ulteriormente espanso o a trovare una soluzione. Nel primo caso si riparte dai nodi al livello precedente di profondità e così via. Nella ricerca in ampiezza invece si espandono sempre i nodi a profondità minore che siano stati generati ma non ancora espansi.  In questo caso si espande prima la radice dell'albero, poi tutti i suoi figli, poi tutti i figli dei figli e così via.  In Figura \ref{prof} è rappresentato l’ordine di espansione dei nodi di un albero secondo la strategia in profondità, e in Figura \ref{amp} secondo la strategia in ampiezza.
      

			Lo spazio di ricerca in realtà può strutturarsi anche in un “grafo”, un albero nel quale un nodo può essere raggiunto seguendo percorsi multipli. In questo caso l’esplorazione può capitare in un nodo già espanso in precendenza. È importante accorgersene per non ripetere operazioni già svolte, e non complicare un problema altrimenti risolvibile. Perciò esistono algoritmi specifici che svolgono la ricerca in grafi anziché in alberi.
      
			Le strategie viste precedentemente sono dette non informate, in quanto non utilizzano criteri per scegliere quale nodo espandere quando ve ne siano più di uno alla stessa profondità. Strategie più efficienti sono invece quelle informate, che, per scegliere, si basano su una conoscenza specifica del problema. Per esempio, la strategia best-first usa una funzione di valutazione $f(n)$ che, dato un nodo $n$, restituisce un valore numerico; per primi sceglie i nodi con il valore $f(n)$ più basso. 
      
			Nella strategia best-first “golosa”, $f(n)$ viene scelta uguale ad una funzione $h(n)$ che, dato un nodo $n$, restituisce una stima del costo del cammino più economico da $n$ ad un nodo finale. $h(n)$ viene chiamata una funzione euristica, in quanto fornisce esclusivamente una stima, per difetto o per eccesso, non un valore certo.
      
			Nella strategia A* (si legge “A star”) la funzione $f(n)$ è data dalla somma di due funzioni, $g(n)$ e $h(n)$. $h(n)$ è una funzione euristica come nel caso precedente, mentre $g(n)$ indica il costo del percorso dallo stato iniziale al nodo $n$. Quindi $f(n)$ in questo caso indica il costo stimato della soluzione più economica che passa attraverso $n$. È possibile dimostrare che, se $h(n)$ rispetta una certa condizione, la strategia A* è completa e ottimale; cioè, se esiste una soluzione, A* la trova, e al costo minimo. La condizione da rispettare è che $h(n)$ sia “ottimistica”, ovvero che non valga mai più del costo reale del cammino più economico da $n$ ad un nodo finale.
      
\begin{figure}[t]
\centering
\includegraphics[width=.55\textwidth]{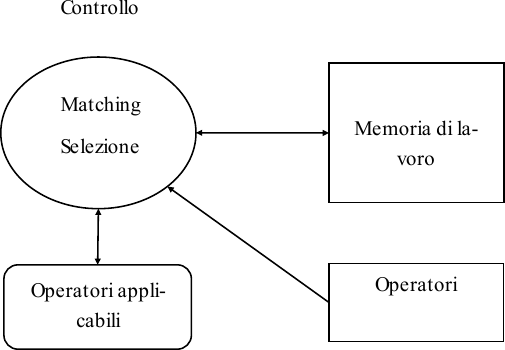}
\caption{Architettura di un sistema di intelligenza artificiale.\label{arch}}
	\end{figure}

			Tutti gli algoritmi visti finora presentano un’architettura comune \cite{mel}: essi hanno una memoria di lavoro per archiviare i risultati parziali, una serie di operatori associabili alle varie situazioni, e una strategia (o controllo), per stabilire quali di questi operatori siano applicabili, sceglierli (se più di uno) e applicarli. Questa architettura, condivisa da molti sistemi di IA, è descritta in Figura \ref{arch}.

      Il tipo di ricerca considerata finora è una ricerca “in avanti” (forward o data-driven), che parte dallo stato iniziale e termina in uno stato finale. È possibile anche una ricerca che proceda in direzione opposta, “all’indietro” (backward o goal-driven), cioè partendo dallo stato finale (o goal) e, applicando gli operatori in senso inverso, arrivando ad uno stato iniziale. In genere, conviene utilizzare la ricerca in avanti quando lo stato iniziale sia uno solo, o pochi, e gli stati finali possano essere molti e  il fattore di ramificazione sia basso vicino allo stato iniziale; mentre è preferibile la ricerca all’indietro quando lo stato finale sia uno solo o pochi, e gli stati iniziali possano essere molti e  il fattore di ramificazione sia basso vicino allo stato finale. Esiste infine un terzo tipo di ricerca, che si chiama “bidirezionale” o “mista”, in cui si procede contemporaneamente in avanti a partire dallo stato iniziale e all’indietro a partire dal goal, terminando quando si incontra lo stesso stato nelle due direzioni. Questo tipo di ricerca ha il vantaggio di restituire due alberi finali, costruiti a partire dallo stato iniziale e da quello finale, che possono avere profondità pari alla metà della profondità di un singolo albero finale costruito invece a partire dallo stato inziale o dallo stato finale. Dato che la dimensione dell’albero cresce in maniera esponenziale con la profondità, la somma delle dimensioni dei due alberi può essere molto più piccola della dimensione dell’albero singolo.
      
			L’approccio alla soluzione dei problemi trattato finora contempla che l’agente intelligente possa prevedere gli effetti delle sue azioni. Il presupposto decade nel caso ci sia più di un agente, umano o artificiale, nel sistema. In particolare, gli altri agenti potrebbero avere obiettivi in conflitto con quelli dell’agente considerato. Si parla allora di “giochi”.
      
      Nei giochi di cui ci occuperemo esistono due giocatori che effettuano mosse a turno, l’effetto delle azioni è deterministico, e la somma delle utilità dei due giocatori al termine del gioco è sempre 0, quindi se uno vince (utilità +1) necessariamente l’altro perde (utilità -1). Inoltre sono giochi in cui si ha informazione perfetta: l’effetto delle mosse è direttamente visibile. 
      
			Esempi sono gli scacchi, la dama, l’othello e il go. Giochi a informazione imperfetta sono invece quelli con le carte, nei quali il giocatore non conosce ciò che ha in mano l’avversario. Programmare un computer per giocare a scacchi è stato uno dei primi obiettivi dell’IA, e già nel 1950 esistevano software del genere. I progressi raggiunti sono stati tali che l’11 maggio del 1997, per la prima volta nella storia, un computer ha battuto in un torneo il campione del mondo di scacchi Garry Kasparov, per 3,5 a 2,5. Il computer era Deep Blue della IBM. Per quanto riguarda la dama e l’othello, le macchine hanno ormai distanziato gli esseri umani, mentre nel backgammon, come negli scacchi, lasciano loro qualche chanche in più. Ma l’unico gioco che ancora resiste davvero è il go, nel quale i computer hanno prestazioni da dilettante.
      
			Giocare può essere visto come un problema di ricerca in cui gli stati sono rappresentati dalle configurazioni dei pezzi sul tavolo da gioco, lo stato iniziale è rappresentato dalla configurazione di partenza dei pezzi, e le mosse possibili per ciascun giocatore sono quelle concesse dalle regole del gioco stesso. A partire da questi dati, si può costruire dalla configurazione iniziale un albero, in questo modo: i figli della radice sono le configurazioni che risultano da una mossa del giocatore che muove per primo, i figli dei figli sono le configurazioni che risultano da una mossa dell’altro giocatore, e così via. Quindi i nodi ai livelli dispari riguardano il primo giocatore, quelli ai livelli pari il secondo. L’albero di ricerca può essere ramificato fino a che non si raggiunga una configurazione finale, nella quale non sia più possibile muovere.
      
			Nel caso dei giochi, è poi necessaria una funzione di utilità che assegna agli stati finali un valore numerico rispecchiante il risultato del gioco stesso: negli scacchi, ad esempio, i valori possono essere +1, -1 o 0, rispettivamente nel caso in cui abbia vinto il primo giocatore, il secondo oppure che la partita sia terminata con un pareggio. In altri giochi i punteggi potrebbero essere diversi; ad esempio, nel backgammon possono andare da +192 a -192. Valori alti dell’utilità sono buoni per il primo giocatore, che chiameremo quindi “max”, mentre valori bassi dell’utilità sono buoni per il secondo giocatore, che chiameremo quindi “min”.
      
			Supponiamo di essere il giocatore max: come selezioniamo la mossa da eseguire in una data configurazione? La soluzione consiste nel costruire l’albero che ha quella configurazione come radice, fino ad arrivare alle configurazioni finali, assegnando poi a ognuna di esse il valore di utilità tramite la funzione di utilità. Dopodichè si assegna un ulteriore valore, che prende il nome di minimax, a tutti i nodi non finali partendo dal basso e procedendo verso l'alto: se al livello del nodo tocca muovere a min, il minimax del nodo è il minimo tra i valori di utilità dei figli; se al livello del nodo tocca muovere a max, il minimax del nodo è il massimo tra i valori di utilità dei figli. Si ripete questo processo ricorsivamente, propagando così i valori di utilità verso la radice.  In questo modo il minimax di un nodo rappresenta l'utilità (per max) del nodo, supponendo che entrambi i giocatori giochino sempre al meglio.  Questo equivale a supporre che min e max facciano sempre la mossa che gli conviene di più, ovvero quella che li conduce all'utilità più alta.  Procedendo a ritroso,  si arriva così ad assegnare un valore minimax a tutti i figli della radice. Il giocatore max eseguirà la mossa che porta alla configurazione con il minimax più alto.
      
			Purtroppo lo spazio di ricerca dei giochi, esclusi i più semplici come il gioco del tris, è enorme e non è quindi possibile costruire l’albero fino agli stati finali. Negli scacchi, è stato stimato che il numero di configurazioni possibili sia dell’ordine di $10^{120}$. In questi casi si costruisce l’albero fino ad una profondità fissata $m$: le foglie dell’albero allora non saranno necessariamente configurazioni finali, e quindi, in luogo dell’utilità, si assegna loro una stima dell’utilità, del tutto simila alla funzione euristica vista in precedenza, ottenuta sulla base della configurazione stessa tramite una funzione di valutazione. Questa funzione deve essere fortemente correlata con le possibilità del giocatore di vincere partendo da quella configurazione. Inoltre dev’essere veloce da calcolare, perché la computazione va eseguita molte volte. Nel caso degli scacchi, un esempio di funzione di valutazione è dato dalla somma dei valori materiali di ciascun pezzo di max meno la somma dei valori materali di ciascun pezzo di min, dove il valore materiale di un pezzo è definito dai testi di scacchi e vale ad esempio 1 per un pedone, 3 per un alfiere o un cavallo, 5 per la torre e così via. Ovviamente le funzioni reali sono più complicate.
      
			Per dare un’idea dell’importanza delle funzioni di valutazione: Deep Blue era in grado di esplorare circa 200 milioni ($2\cdot 10^8$) di posizioni al secondo \cite{but}. L’esplorazione dell’intero spazio di ricerca avrebbe dunque richiesto a Deep Blue $5\cdot 10^{111}$ secondi, ovvero 1095 miliardi di anni! Quindi la porzione dello spazio di ricerca effettivamente esplorata è stata piccola, e sono state usate complesse e articolate funzioni di valutazione. Al confronto con Kasparov, però, le funzioni di valutazione di Deep Blue sono semplici, se si pensa che lo scacchista russo, pur elaborando, secondo una stima, solo 3 posizioni al secondo, è riuscito a dare del filo da torcere al computer! Per questo si dice che Deep Blue abbia usato per vincere la forze bruta, piuttosto che una intelligenza raffinata.

\subsection{Ragionamento automatico}
Per ragionamento automatico s’intende l’utilizzo di conoscenza al fine di inferire nuova conoscenza. A questo scopo è necessario rappresentare la conoscenza in un formato memorizzabile da un calcolatore, e utilizzabile per effettuare inferenze. Questi requisiti restringono il formato di rappresentazione a linguaggi formali, ovvero linguaggi con una sintassi e una semantica definiti in maniera precisa.
      
			Uno dei linguaggi formali maggiormente studiati è quello della logica. Essa ha le sue origini nella filosofia e nella matematica greca antica. Il padre fondatore della logica come disciplina autonoma può essere considerato Aristotele (384-321 a.C. circa), mentre Crisippo di Soli (280-205 a.C. circa), della scuola stoica, definì i connettivi logici, gli assiomi e le regole fondamentali della logica proposizionale.
      
			La nascita della moderna logica matematica può essere fatta risalire a George Boole (1815-1864), che nel 1847 pubblicò un metodo per descrivere la teoria dei sillogismi aristotelici e la logica proposizionale sotto forma di equazioni algebriche, e propose un procedimento meccanico per la loro soluzione. Gottlob Frege (1848-1925) fu il primo a sviluppare un sistema di assiomi e regole per la logica del primo ordine, superando così i limiti imposti dai sillogismi e dalla logica proposizionale.
      
			Nel 1965 John Robinsono pubblicò il metodo di risoluzione, che consente di automatizzare in maniera efficiente l’inferenza deduttiva nella logica del primo ordine. Su questo metodo è basata la programmazione logica, e il linguaggio Prolog (PROgramming in LOGic) in particolare, le cui basi furono gettate da alcuni ricercatori delle Università di Edimburgo e Marsiglia nei primi anni ‘70. Soprattutto Robert Kowalski, a Edimburgo, si occupò di definire i fondamenti teorici della programmazione logica, e propose una interpretazione procedurale delle formule logiche che consente di ridurre il processo di dimostrazione di un teorema ad un processo di computazione su un tradizionale elaboratore. Alain  Colmerauer a Marsiglia fu il primo a realizzare, nel 1972, un interprete per il linguaggio Prolog.
      
			Vediamo alcuni rudimenti della programmazione logica, cominciando dalla logica proposizionale. In essa, le unità elementari sono le proposizioni atomiche, cioè affermazioni non ulteriormente scomponibili che possono essere vere o false. Le proposizioni arbitrarie o formule proposizionali sono ottenute dalle formule atomiche, combinandole mediante i connettivi logici: negazione, congiunzione, disgiunzione e implicazione.
      
			Esempi di proposizioni atomiche sono le affermazioni “l’erba è bagnata” e “oggi piove”. Un esempio di formula proposizionale è l’espressione “l’erba è bagnata se oggi piove” che è una implicazione. Nella logica proposizionale, si utilizzano lettere per indicare le proposizioni e simboli per i connettivi. Ad esempio, $b$ per “l’erba è bagnata”, $p$ per “oggi piove” e $\leftarrow$ per l'implicazione. La formula proposizionale composta, mostrata sopra, si può dunque scrivere come
      $$b\leftarrow p$$
      In programmazione logica, si possono usare sequenze di lettere per rappresentare le proposizioni, e si usa il simbolo \verb|:-| per l’implicazione logica. Si può quindi rappresentare la formula proposizionale composta come:
\begin{verbatim}
erba_bagnata :- oggi_piove.
\end{verbatim}
      In questo caso la sequenza di lettere ci serve per ricordare il significato delle proposizioni.
      
      Inferenze nella logica proposizionale possono essere compiute mediante regole di inferenza. La più semplice è il modus ponens che da $p$ e $b\leftarrow p$ deriva $b$. Questa regola rappresenta un passo elementare di deduzione, ed è corretta, nel senso che se sono vere le premesse allora è vera anche la conclusione. Quindi se è vero che “oggi piove” ed è vero che “l’erba è bagnata se oggi piove”, allora si può concludere che “l’erba è bagnata”. In programmazione logica si scrive il seguente programma:
\begin{verbatim}
oggi_piove.
erba_bagnata :- oggi_piove.
\end{verbatim}
e si interroga il sistema sulla verità di \verb|erba_bagnata| scrivendo \verb|erba_bagnata.| sulla riga di comando di un interprete Prolog. 
      Si ottiene così la riposta:
\begin{verbatim}
yes
\end{verbatim}
che significa che \verb|erba_bagnata| è vero.

      La verità di \verb|erba_bagnata| può poi essere utilizzata per derivare altre proposizioni, applicando ripetutamente il modus ponens. 
      
      Nella logica del primo ordine le formule atomiche differiscono da quelle della logica proposizionale perché possono avere uno o più argomenti. Gli argomenti sono termini, ovvero rappresentazioni di un individuo del dominio del discorso. Nei casi più semplici, i termini sono variabili se indicano un individuo imprecisato, o costanti se determinano un individuo specifico. Nel seguito utilizzeremo la convenzione del Prolog di usare parole che iniziano con lettere minuscole per indicare costanti, e parole che iniziano con lettere maiuscole per indicare variabili. Le proposizioni diventano predicati ed esprimono proprietà dei loro argomenti. Ad esempio $p(a)$ esprime il fatto che “l’individuo $a$ è $p$” o “l’individuo $a$ ha la proprietà $p$” e $q(a,b)$ esprime il fatto che “la coppia di individui $a$ e $b$ è $q$” ovvero che “la coppia $a$ e $b$ ha la proprietà $q$”, ovvero che “$a$ e $b$ sono legati dalla relazione $q$”. Esempi di formula atomica della logica proposizionale sono $\textit{uomo(socrate)}$, che significa che “Socrate è un uomo”, e $\textit{padre(paolo,pietro)}$, che indica che Paolo e Pietro sono legati dalla relazione padre, ovvero che Paolo è il padre di Pietro.

      Le formule atomiche nella logica del primo ordine possono essere combinate con gli stessi connettivi logici della logica proposizionale. In più si possono ottenere nuove formule utilizzando i quantificatori: quello esistenziale che, data una variabile $X$ e una formula $\alpha$, produce la formula $\exists X \alpha$ e quello universale che, data una formula $\alpha$, produce la formula $\forall X \alpha$. I quantificatori hanno senso se $\alpha$ contiene $X$: in questo caso la formula $\exists X \alpha$ significa che esiste un individuo $i$ del dominio del discorso che, sostituito a $X$, rende la formula $\alpha$ vera, mentre la formula $\forall X \alpha$ significa che, per ogni individuo $i$ del dominio del discorso, sostituendo $i$ ad $X$ in $\alpha$ si ottiene una formula vera.

      Ad esempio, la formula
$$\forall       X (mortale(X) \leftarrow uomo(X))$$
significa che, per qualunque individuo $X$ del dominio del discorso, se $X$ è uomo allora $X$ è mortale. In Prolog si rappresenta la formula in questo modo
\begin{verbatim}
mortale(X) :- uomo(X).
\end{verbatim}
sottintendendo il quantificatore universale.

      Questa formula prende il nome di clausola o regola. Un programma Prolog può comprendere poi un altro tipo di formule chiamate fatti, che non contengono il simbolo di implicazione. Ad esempio, potremmo aggiungere al programma precedente il fatto \verb|uomo(socrate).| e interrogare poi il programma Prolog chiedendo se Socrate è mortale. Per farlo, dovremo scrivere l’interrogazione \verb|mortale(socrate).| sulla linea di comando dell’interprete Prolog. Il Prolog rispondera con \verb|yes|, in quanto \verb|mortale(socrate)| è conseguenza logica del programma. In questo caso il modus ponens non è più sufficiente, occorre la risoluzione, la regola di inferenza proposta da John Robinson.

      Al programma precedente potremmo anche chiedere se c’è qualcuno che è mortale, scrivendo \verb|mortale(M).| sulla riga di comando dell’interprete Prolog. L’interprete non risponde più semplicemente con un sì o con un no ma con una istanziazione di \verb|M| che rende la formula vera. Nel caso precedente risponderà con \verb|M=socrate| e si fermerà per chiedere se vogliamo altre istanziazioni per \verb|M|. Se le chiediamo, in questo caso risponderà \verb|no|, ma se aggiungessimo un altro fatto al programma, ad esempio \verb|uomo(platone).|, il sistema fornirà, oltre alla risposta \verb|M=socrate|, anche la risposta \verb|M=platone|.

      Le regole in Prolog possono poi contenere, nel lato destro di una implicazione, una congiunzione di formule atomiche. La congiunzione in Prolog è rappresentata dalla virgola. Ad esempio, la clausola:
\begin{verbatim}
padre(X,Y) :- genitore(X,Y),maschio(X).
\end{verbatim}
significa che \verb|X| è padre di \verb|Y| se \verb|X| è genitore di \verb|Y| e \verb|X| è maschio.
      
      Il lato destro può anche contenere variabili non presenti nel lato sinistro, ad esempio:
\begin{verbatim}
nonno(X,Y) :-	padre(X,Z),genitore(Z,Y),
\end{verbatim}
in questo caso le variabili presenti solo nel lato destro sono quantificate esistenzialmente con ambito di quantificazione il solo lato destro. Quindi la regola precedente si può interpretare come: \verb|X| è nonno di \verb|Y| se esiste un \verb|Z| tale che \verb|X| è padre di \verb|Z| e \verb|Z| è genitore di \verb|Y|.
    
			Si noti che nel lato destro può comparire anche il predicato della formula nel lato sinistro, ad esempio:
\begin{verbatim}
antenato(X,Y) :- padre(X,Z), antenato(Z,Y).
\end{verbatim}
      Questa regola si interpreta in questo modo: “\verb|X| è antenato di \verb|Y| se esiste un \verb|Z| tale che \verb|X| è padre di \verb|Z| e \verb|Z| è antenato di \verb|Y|”. Si parla di definizione ricorsiva.
    
			Il Prolog, quindi, non è altro che un dimostratore di teoremi, quelli scritti sulla riga di comando dell’interprete. L’architettura di un interprete Prolog è una istanziazione di quella generale di un sistema di IA: ha una memoria di lavoro, dove risiedono i risultati parziali, ha una serie di operatori che sono le regole e i fatti del programma Prolog, e usa una strategia o controllo per stabilire quali regole o fatti sono applicabili, decidere quali applicare ed applicarli effettivamente.
			
			L’insieme della memoria di lavoro e della strategia di controllo prende il nome di motore inferenziale, mentre l’insieme delle regole e dei fatti viene chiamato base di conoscenza. In particolare, il Prolog compie una ricerca partendo dal goal o teorema da dimostrare, applicando via via le regole o i fatti del programma. nel modo seguente: si cercano tutte le regole o fatti tali che il loro lato sinistro “unifichi” con il goal o con una parte di esso (“unificare” significa rendere uguali il goal con il lato sinistro, attraverso l’istanziazione di variabili) Ad esempio, nel caso del goal \verb|mortale(socrate)|, esso unifica con il lato sinistro della regola \verb|mortale(X):-uomo(X).| istanziando \verb|X| a \verb|socrate|. Tra le regole che hanno il lato sinistro che unifica, se ne sceglie una (ma non si dimenticano le altre) e si sostituisce il goal con l’elenco di sottogoal nel lato destro della regola. Si procede in questo modo fino a che non si arriva ad un goal vuoto (e allora si termina con successo restituendo le sostituzioni nel goal iniziale), oppure fino a che non ci siano più regole o fatti applicabili (nel qual caso si retrocede al punto di scelta della regola e se ne sceglie un’altra). Il Prolog esplora quindi un albero in modalità all’indietro. La sua memoria di lavoro consiste del goal corrente più tutti i punti di scelta lasciati aperti, e la strategia seguita è una strategia in profondità.
      
			Mediante il Prolog si possono costruire sistemi esperti, anche detti sistemi basati sulla conoscenza, caratterizzati dall’essere dotati di una conoscenza che riguarda uno specifico campo, e dall’essere in grado di risolverne i problemi relativi, che un esperto sarebbe in grado di risolvere. Nel caso di sistemi esperti basati sul Prolog, la conoscenza viene espressa sotto forma di programma logico, e il metodo di inferenza del Prolog è usato per trovare soluzioni al problema.
      
			L’uso del Prolog non è l’unico approccio per la realizzazione di sistemi esperti: si possono utilizzare anche sistemi a regole di produzione. Essi condividono la struttura generale dei sistemi di IA: in questo caso la base di conoscenza contiene regole della forma “conseguente se antecedente” simili a quelle del Prolog, ma ove nel conseguente non sia presente una formula da derivare ma una o più azioni sulla memoria di lavoro. Le azioni possibili sono due: inserimento di un fatto oppure rimozione di un fatto. Tipicamente le regole esprimono conoscenza generale sul dominio mentre la conoscenza specifica sul caso in esame è espressa da fatti. La strategia di ricerca può essere sia all’indietro, come in Prolog (partendo dal goal e applicando le regole a ritroso fino a ottenere una memoria di lavoro con i fatti che descrivono il problema), sia in avanti (si pongono i fatti che descrivono il problema nella memoria di lavoro, si cercano le regole che presentano l’antecedente soddisfatto, se ne sceglie una e si applicano le azioni contenute nel conseguente, aggiungendo o rimuovendo un fatto dalla memoria stessa). La ricerca termina quando il goal appare nella memoria di lavoro. Per stabilire quali regole sono applicabili nei due casi si utilizzano algoritmi di unificazione o che mettano in corrispondenza tra loro (“matching”) formule atomiche. A differenza del Prolog, in un sistema a regole di produzione quando una regola viene scelta non si considerano più le possibili alternative.
      
			Alcuni sistemi a regole di produzione includono inoltre la gestione dell’incertezza, cioé sono in grado di assegnare alle proprie conclusioni un livello di confidenza. Attualmente però non è stato ancora raggiunto un consenso tra i ricercatori su come utilizzare i livelli di confidenza nell’inferenza.
      
			Sistemi esperti sono stati sviluppati in molti domini. Il primo e forse il più noto è Mycin creato negli anni ‘70 all’Università di Stanford da Edward Shortliffe. Aveva come obiettivo quello di diagnosticare malattie infettive del sangue e raccomandare antibiotici, con un dosaggio adattato al peso del paziente. 
			Il sistema offriva buone prestazioni, ma non venne mai utilizzato, per problemi legali.
      
			Alcuni settori in cui possono essere impiegati sistemi esperti sono: 
			\begin{enumerate}[a)]
			\item diagnosi, nella quale si cerca di individuare una malattia di un essere umano o il malfunzionamento di un macchinario sulla base dei sintomi, ovvero delle manifestazioni osservabili della malattia o malfunzionamento; 
 \item monitoraggio, in cui l’obiettivo è di mantenere sotto controllo un processo, raccogliendo informazioni e effettuando stime sul suo andamento; 
      \item pianificazione, in cui l’obiettivo è quello di raggiungere un certo obiettivo con le risorse di cui si dispone; 
      \item interpretazione di informazioni e segnali, in cui si vuole individuare l’occorrenza di particolari situazioni di interesse nei dati in ingresso.
      \end{enumerate}
			Lo sviluppo di un sistema esperto richiede la scrittura delle regole generali sul dominio, che devono essere raccolte intervistando un esperto del dominio stesso. Questo processo, conosciuto con il nome di estrazione di conoscenza, è risultato essere estremamanente lungo e difficile. Al fine di automatizzarlo, è possibile usare l’apprendimento automatico, discusso nella prossima sezione.
      
			Ci si potrà chiedere perchè siano stati realizzati solo sistemi esperti in domini ristretti e non si siano utilizzate queste tecniche per sviluppare un sistema in grado di essere applicato in ogni possibile situazione, comprese quelle cosiddette di senso comune, ovvero relative a ragionamenti che ciascuno di noi fa quotidianamente. La ragione è l’eccessiva complessità della fase di stesura delle regole. Recentemente un’azienda statunitense, la Cycorp, sta tentando nuovamente di codificare il senso comune in un sistema esperto.

\subsection{Apprendimento automatico}
Simon nel 1984 ha dato la seguente definizione di apprendimento \cite{sim}: “L’apprendimento consiste di cambiamenti del sistema che siano adattativi, nel senso che mettono in grado il sistema di svolgere lo stesso compito o compiti estratti dalla medesima popolazione in maniera più efficace ed efficiente la prossima volta”. Di sicuro, al fine di realizzare macchine che possano dirsi intelligenti, è necessario dotarle della capacità di estendere la propria conoscenza e le proprie abilità in modo autonomo.
      
			I due impieghi principali dell’apprendimento automatico sono l’estrazione di conoscenza e il miglioramento delle prestazioni di una macchina. La conoscenza estratta può poi essere utilizzata da una macchina come base di conoscenza di un sistema esperto, oppure dagli esseri umani, ad esempio nel caso della scoperta di nuove teorie scientifiche. Il miglioramento delle prestazioni di una macchina si ha ad esempio quando si incrementano le capacità percettive e motrici di un robot.
      
			Le tecniche di apprendimento si possono suddividere, come le tecniche di IA in generale, in simboliche e subsimboliche. 
      
      La tecnica di apprendimento simbolico più interessante è l’apprendimento induttivo: il sistema parte dai fatti e dalle osservazioni derivanti da un insegnante o dall’ambiente circostante, e li generalizza ottenendo conoscenza che, auspicabilmente, sia valida anche per casi non ancora osservati (induzione).
      
      Nell’apprendimento induttivo da esempi, l’insegnante fornisce un insieme di esempi e controesempi di un concetto, e l’obiettivo è quello di inferire una descrizione del concetto stesso. Un esempio è composto da una descrizione di una istanza del dominio del discorso e da una etichetta; quest’ultima può essere + se l’istanza appartiene al concetto da apprendere, o – se l’istanza non gli appartiene (controesempio). Un concetto, quindi, non è altro che un sottoinsieme dell’insieme di tutte le possibili istanze del domino del discorso, o universo. L’insieme di esempi e controesempi forniti dall’insegnante prende il nome di training set. La descrizione del concetto che si vuole apprendere deve essere tale da potersi usare per decidere se una nuova istanza, non appartenente al training set, appartenga o meno al concetto. La descrizione del concetto deve essere quindi un algoritmo che, data in ingresso una descrizione dell’istanza, restituisca in uscita +, se l’istanza appartiene al concetto, o -, se l’istanza non appartiene al concetto. Nel primo caso si parla di esempio appartenente alla classe positiva e nel secondo di esempio appartenente alla classe negativa.
      
			I sistemi di apprendimento induttivo da esempi possono essere classificati in base al linguaggio con il quale è possibile descrivere le istanze e i concetti, che sono principalmente due: linguaggi attributo valore e linguaggi del primo ordine. Ai primi corrispondono, come linguaggi di descrizione dei concetti, gli alberi di decisione e le regole di produzione, mentre ai secondi corrispondono descrizioni dei concetti anch’esse del primo ordine.
      
			I linguaggi di descrizione delle istanze di tipo attributo valore descrivono ciascuna istanza per mezzo dei valori assunti da un insieme finito di attributi uguali per tutte le istanze. Ad esempio; si supponga che le istanze siano giornate e che il concetto che vogliamo imparare sia “giornata adatta per giocare a golf” \cite{mit}. Supponiamo di aver individuato 4 attributi che pensiamo siano rilevanti rispetto al concetto da imparare: Tempo, Temperatura, Umidità e Vento.
      
			Tempo può assumere solo tre valori: soleggiato, coperto e pioggia; Vento può assumere solo due valori: vero (v) e falso (f), nel caso sia presente o non sia presente vento. Temperatura e Umidità invece sono attributi continui, possono assumere quindi valori da un intervallo dell’insieme dei reali. Temperatura è espressa in gradi Fahrenheit e Umidità è espressa in percentuale. Nel caso di istanze descritte da un linguaggio attributo valore possiamo rappresentare il training set per mezzo di una tabella con una colonna per attributo e una riga per istanza. La tabella inoltre avrà una colonna aggiuntiva che conterrà la classe di appartenenza dell’istanza, ovvero + se l’istanza appartiene al concetto e – se l’istanza non appartiene al concetto.
			Un possibile training set per il problema del golf è rappresentato in Tabella \ref{ist}.
\begin{table}
\centering
\begin{tabular}{|l|l|l|l|l|}
\hline
Tempo&Temperatura&Umidità&Vento&Classe\\\hline
soleggiato&75&70&v&+\\\hline
soleggiato&80&90&v&-\\\hline
soleggiato&85&85&f&-\\\hline
soleggiato&72&95&f&-\\\hline
soleggiato&69&70&f&+\\\hline
coperto&72&90&v&+\\\hline
coperto&83&78&f&+\\\hline
coperto&64&65&v&+\\\hline
coperto&81&75&f&+\\\hline
pioggia&71&80&v&-\\\hline
pioggia&65&70&v&-\\\hline
pioggia&75&80&f&+\\\hline
pioggia&68&80&f&+\\\hline
pioggia&70&96&f&+\\\hline
\end{tabular}
\caption{Istanze di giornate (da \cite{mit}).\label{ist}}
\end{table}

\begin{figure}[t]
\centering
\includegraphics[width=.95\textwidth]{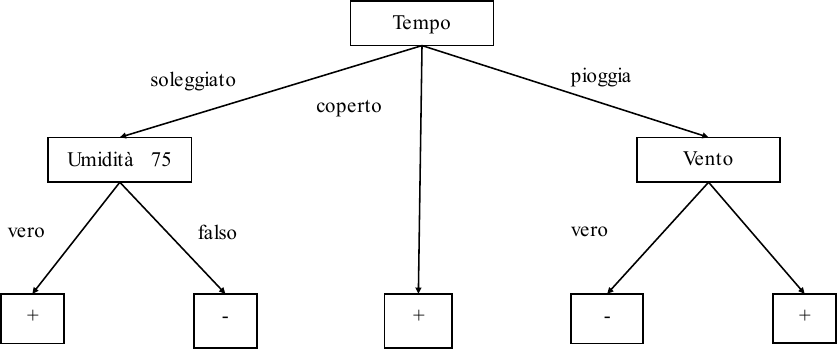}
\caption{Albero di decisione per il problema del golf.\label{albdec}}
	\end{figure}

      Gli alberi di decisione sono un linguaggio di descrizione dei concetti che può essere usato quando le istanze sono descritte da linguaggi attributo valore. Negli alberi di decisione ogni nodo corrisponde ad un test su un attributo, e ciascun ramo che parte dal nodo è etichettato con il risultato del test. Il test su un attributo discreto consiste in un test di uguaglianza e i possibili risultati sono i possibili valori discreti dell’attributo. Ad esempio, nel caso di Tempo i possibili risultati sono soleggiato, coperto e pioggia. Il test su un attributo continuo consiste nel confronto dell’attributo con una soglia, e i possibili risultati sono due: attributo minore o uguale della soglia o attributo maggiore della soglia. Le foglie sono etichettate con + o -. L’albero può essere usato per classificare un nuova istanza in questo modo: si parte dalla radice e si considera il test nella radice. In base al valore per l’istanza dell’attributo usato dal test, si sceglie il ramo lungo cui scendere. Si ripete il procedimento fino a che non si arriva ad una foglia: l’etichetta della foglia indica l’appartenenza o meno dell’istanza al concetto. Un albero di decisione per il problema del golf è rappresentato in Figura \ref{albdec}.
			
      Le regole di produzione sono un altro linguaggio di descrizioni dei concetti che può essere usato quando le istanze sono descritte da linguaggi attributo valore. 
      
			Come nei sistemi a regole di produzione visti in precedenza, le regole di produzione nell’apprendimento automatico hanno la forma “conseguente se antecedente” con l’unica differenza che nell’apprendimento automatico il conseguente ha una forma fissa: è del tipo “Classe=valore”. Nell’antecedente possono comparire confronti di un attributo con un valore: nel caso di attributi discreti il confronto è fatto mediante l’uguaglianza, mentre nel caso di attributi continui il confronto è fatto mediante una disuguaglianza. Le regole sono utilizzate per classificare una nuova istanza restituendo la classe indicata dalla prima regole il cui antecedente è verificato nell’istanza. Vediamo un esempio di insieme di regole di produzione per il problema del golf:
      
			Classe = + $\leftarrow$ Tempo = coperto

      Classe = + $\leftarrow$ Tempo = pioggia and Vento = falso

      Classe = - $\leftarrow$ Tempo = soleggiato and Umidità > 75

      Classe = - $\leftarrow$ Tempo = pioggia and Vento = vero

      Classe = + 
      
			L’ultima regola viene chiamata regola di default perchè indica la classe in delle istanze quando nessuna altra regola è applicabile.
      
			Si noti che l’insieme di regole di produzione restituite da un sistema di apprendimento automatico costituisce la base di conoscenza di un sistema a regole di produzione, sebbene particolarmente semplice in quanto non presenta concatenazione tra le regole.
      
			Esistono numerosi sistemi in grado di apprendere alberi di decisione: i più noti sono Cart, sviluppato da Breiman, Friedman, Olshen e Stone, e ID3, sviluppato da Quinlan, entrambi nei primi anni ‘80. Tra i sistemi in grado di apprendere regole di produzione ricordiamo CN2 di Clark e Niblett sviluappato nella seconda metà degli anni ‘80.
      
			I linguaggi attributo valore hanno alcune limitazioni: essi non sono adatti a descrivere istanze costituite di sottoparti e aventi relazioni tra le sottoparti. Ad esempio, si supponga che le istanze del dominio del discorso siano famiglie. Le famiglie hanno un numero variabile di componenti, quindi per descriverle attraverso un insieme fisso di attributi dovremmo prevedere un numero di attributi pari al numero massimo di componenti di una famiglia moltiplicato per il numero di attributi che vogliamo rappresentare per ogni componente. Per le famiglie con un numero non massimo di componenti, ad alcuni attributi si dovrà assegnare un valore di “non significativo”. In questi casi risulta essere più efficace utilizzare la programmazione logica come linguaggio di rappresentazione. Ad esempio, se voglio rappresentare la famiglia Rossi con componenti Giorgio, Stefano e Andrea nella quale Giorgio è il padre di Stefano e Stefano è il padre di Andrea scriverò semplicemente:
\begin{verbatim}      
famiglia(rossi,giorgio).
famiglia(rossi,stefano).	
famiglia(rossi,andrea).
padre(giorgio,stefano).	
padre(stefano,andrea).
\end{verbatim}      
      Il linguaggio di descrizione dei concetti in questo caso è la programmazione logica stessa. Essa risulta essere più espressiva degli alberi di decisione e delle regole di produzione in quanto i programmi logici possono contenere variabili e quantificatori, e le regole possono essere ricorsive. Mediante la programmazione logica è possibile esprimere il concetto di famiglia che contiene un nonno paterno:
\begin{verbatim}
nonno_paterno(Famiglia):-
  famiglia(Famiglia,X), 
  famiglia(Famiglia,Y), 
  famiglia(Famiglia,Z).
  padre(X,Y), 
  padre(Y,Z).
\end{verbatim}
      Con gli alberi di decisione o le regole di produzione non si sarebbe potuto esprimere questo concetto.
    
			L’area di ricerca che si occupa dell’apprendimento di programmi logici prende il nome di Programmazione Logica Induttiva. I due sistemi più noti in grado di apprendere programmi logici sono Progol di Muggleton e Aleph di Srinivasan.
      
			I sistemi di apprendimento, sia da linguaggi attributo valore che da programmi logici, hanno avuto una vasta gamma di applicazioni, che va dalla diagnosi di malattie alla predizione della relazione struttura-attività nella progettazione di medicine, alla predizione della carcinogenicità delle sostanze chimiche. Con l’aumento della quantità di dati che vengono memorizzati ogni giorno dalle aziende e dalle organizzazioni in generale, gli algoritmi di apprendimento sono sempre più importanti in quanto consentono di estrarre da questa massa di dati informazioni nascoste, nuove e potenzialmente utili. Si parla in questo caso di data mining, ovvero di estrazione di conoscenza da dati grezzi.

\section{Tecniche subsimboliche}
Vedremo ora tre tecniche subsimboliche: le reti neurali, gli algoritmi genetici e l’intelligenza degli sciami.
\subsection{Reti neurali}
L’idea di simulare il funzionamento del cervello umano e animale per ottenere comportamenti intelligenti risale a prima della realizzazione del computer, in particolare all’articolo del 1943 di McCulloch e Pitts \cite{mccpit} nel quale si propose un modello matematico del neurone umano e si mostrò come reti composte di tali neuroni artificiali fossero in grado di rappresentare complesse funzioni booleane.

\begin{figure}[t]
\centering
\includegraphics[width=.95\textwidth]{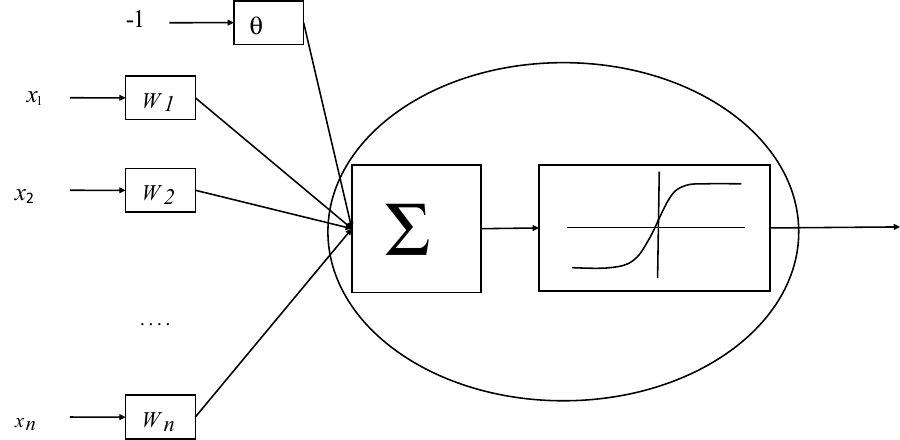}
\caption{Modello del neurone artificiale.\label{ann}}
	\end{figure}

      Il modello del neurone attualmente più diffuso è chiamato neurone sigmoidale ed è costituito da una unità con n ingressi numerici e una uscita numerica. L’uscita è calcolata in funzione degli ingressi nel seguente modo: ciascun ingresso $x_i$ viene moltiplicato per un peso $W_i$, i prodotti di queste moltiplicazioni sono sommati e il risultato viene fornito in ingresso ad una funzione sigmoidale. Un modello di questo tipo di neurone è rappresentato in Figura \ref{ann}, insieme all’aspetto della funzione sigmoidale. Si noti che nella somma c’è un termine costante pari a $-\theta$ che viene assimilato ad un comune ingresso supponendo che l’ingresso sia sempre a -1 e che il peso per quell’ingresso valga i.

      Questo modello si comporta in maniera simile a un neurone naturale: ovvero si “attiva” quando riceve gli ingressi “giusti” e si “disattiva” in corrispondenza di ingressi “sbagliati”. Un neurone è attivo quando la sua uscita è vicina a +1 ed è disattivo quando la sua uscita è vicina a -1. Quali siano gli ingressi giusti o sbagliati è determinato dai valori dei pesi degli ingressi: valori positivi dei pesi fanno sì che i relativi ingressi tendano a portare il neurone verso l’attivazione e valori negativi verso la disattivazione e viceversa nel caso di pesi negativi. I neuroni sono poi collegati tra di loro in reti, quindi l’uscita di un neurone può essere l’ingresso di altri neuroni e la sua attivazione condiziona l’attivazione dei neuroni a valle.

      Il neurone sigmoidale deriva dal percettrone proposto nel 1962 da Rosenblatt: differisce da esso perchè al posto della funzione sigmoidale il percettrone ne ha una a gradino, cioè una funzione che è 0 per i valori minori di 0 e 1 per i valori maggiori o uguali a 0.

      Un singolo neurone sigmoidale è in grado di rappresentare una certa classe di concetti a seconda dei suoi pesi: in particolare, è in grado di rappresentare quei concetti nei quali gli esempi sono separati dai controesempi da un iperpiano (immaginando di considerare lo spazio degli ingressi come uno spazio euclideo). Quindi isolano nello spazio degli ingressi un semispazio, ovvero una regione delimiatata da un iperpiano (nel caso di due ingressi si tratta di una retta). Per rappresentare concetti più complessi è necessario comporre i neuroni in reti.

\begin{figure}[t]
\centering
\includegraphics[width=.6\textwidth]{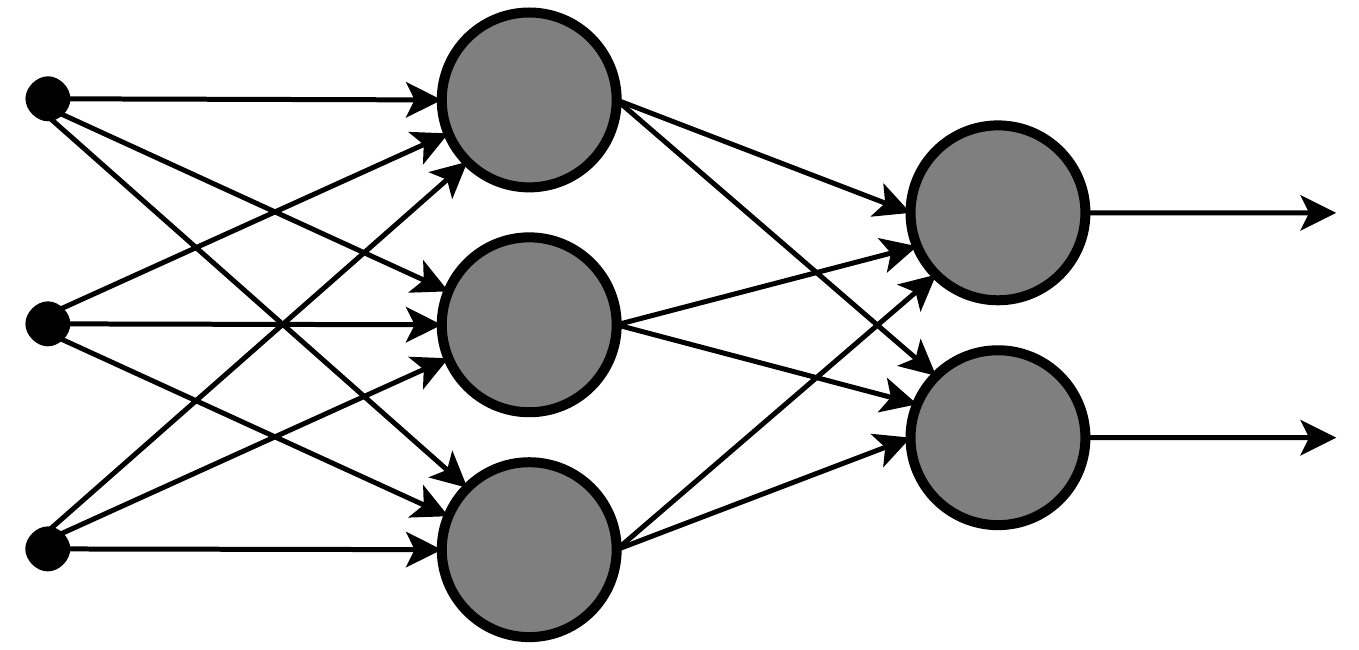}
\caption{Una rete neurale con strati multipli (A Neural network with multiple layers -- Offnfopt -- \href{https://commons.wikimedia.org/wiki/File:Multi-Layer_Neural_Network-Vector-Blank.svg}{wikipedia} -- CC BY-SA 3.0).\label{mlp}}
	\end{figure}

      Le reti più semplici si chiamano reti “in avanti” (“feedforward”) e sono formate da strati di neuroni: gli input sono collegati al primo strato di neuroni, gli output del primo strato di neuroni sono collegati agli input del secondo strato e così via, fino ad arrivare all’ultimo strato i cui output diventano gli output della rete. Gli strati di neuroni dal primo al penultimo sono detti “nascosti”. La Figura \ref{mlp} mostra una rete feedforward con uno strato nascosto.

      Come fare però ad ottenere una rete che identifichi un concetto? A differenza dei sistemi basati sulla conoscenza, nelle reti neurali scegliere i pesi a mano risulterebbe troppo complesso. Per questo si utilizzano algoritmi di apprendimento. Anche in questo caso abbiamo un training set, che contiene un insieme di coppie (ingressi, uscite) con la differenza che gli ingressi sono tutti continui, le uscite possono essere più di una e anch’esse continue, ovvero non sono + o – ma numeri reali. Si cerca in questo caso il valore dei pesi per il quale è minima una certa funzione dell’errore sul training set, ovvero una funzione delle differenze tra le uscite nel training set e quelle della rete quando in ingresso sono forniti i valori presenti nella coppia. La funzione più utilizzata è la somma degli errori al quadrato.

      L’algoritmo più diffuso per apprendere in reti neuronali multistrato prende il nome di “backpropagation” ed è stato proposto da Rumelhart, Hinton e Williams nel 1986. Esso consiste nel calcolare l’errore dello strato d’uscita della rete su ciascun esempio e nel propagarlo all’indietro verso i neuroni degli strati nascosti. Sulla base dell’errore propagato vengono poi aggiornati i pesi dei neuroni. Se dopo aver considerato in questo modo tutti gli esempi l’errore è sceso sotto una soglia prefissata ci si ferma, altrimenti si considerano un’altra volta tutti gli esempi del training set.

Il teorema dell'approssimazione universale \cite{DBLP:journals/mcss/Cybenko89} dice che con una rete dotata di un solo strato nascosto è possibile approssimare una qualsiasi funzione, purché  non si ponga limite al numero di neuroni nello strato 
nascosto. Si è però scoperto che con reti con più di uno strato nascosto è possibile approssimare funzioni utilizzando meno neuroni. Da questa osservazione è nato il settore del “deep learning”: grazie all'incremento della potenza di calcolo dei
computer è stato possibile negli ultimi 30 anni addestrare reti con un numero sempre maggiore di strati nascosti, ovvero reti “profonde” (deep).

Le reti multistrato deep  hanno avuto un grande successo e sono state applicate in molti campi, tra cui la visione artificiale, il riconoscimento del parlato, la traduzione automatica, la bioinformatica, la progettazione di medicine e l'analisi di immagini mediche.

\begin{figure}[t]
\centering
\includegraphics[width=.6\textwidth]{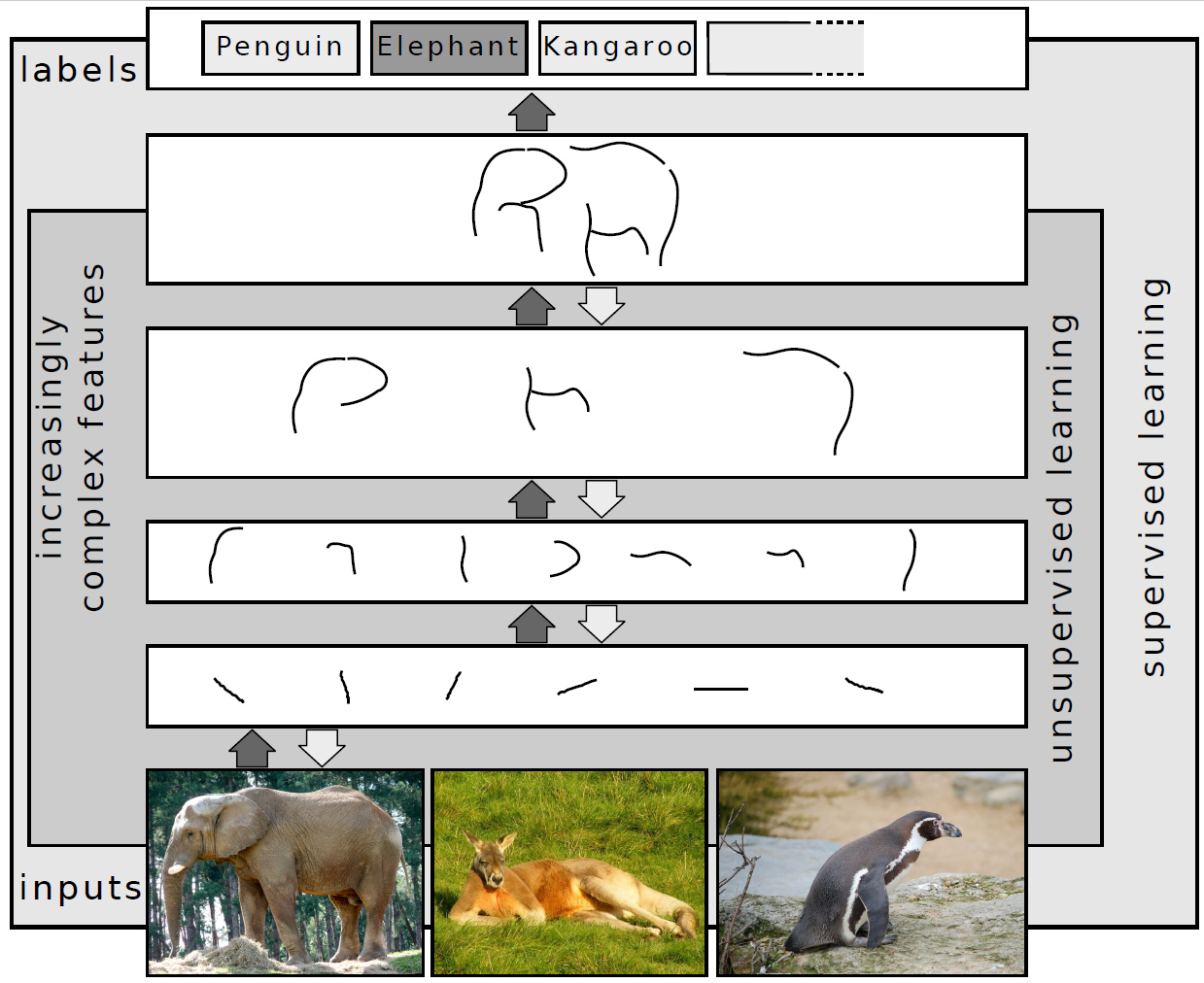}
\caption{Livelli di astrazione multipli (Schematic overview of layer-wise learning -- Sven Behnke -- \href{https://commons.wikimedia.org/wiki/File:Deep_Learning.jpg}{wikipedia} -- CC BY-SA 3.0) .\label{dl}}
	\end{figure}

Nella computer vision ad esempio sono state in grado di riconoscere caratteri scritti a mano, riconoscere oggetti in immagini e video e classificare immagini. Questi risultati sono stati ottenuti utilizzando un particolare tipo di rete
neurale detta convoluzionale. In questa rete alcuni strati applicano all'input un'operazione di convoluzione: un filtro viene fatto scorrere sopra l'input bidimensionale producendo così un'immagini filtrata. Questi filtri, che sono addestrati
insieme ai parametri degli strati tradizionali, identificano  caratteristiche dell'immagine di complessità crescente: ad esempio, un primo strato convoluzionale potrebbe identificare bordi approssimativamente rettilinei nell'immagine di input, un secondo strato
combinazioni di bordi rettilinei (angoli) e così via, fino ad arrivare al punto di identificare quelle caratteristiche dell'immagine che servono a classificarla. Ad esempio, nella Figura \ref{dl} vediamo le caratteristiche che sono estratte ai vari livelli 
da una immagine di input: dal primo strato che identifica porzioni di linea, si arriva all'ultimo strato convoluzionale che identifica in questo caso la silhouette di un elefante e classifica di conseguenza l'immagine
\cite{deep_learn}.

Attualmente le reti convoluzionali deep sono in grado di superare l'uomo in molti compiti di visione artificiale.

\subsection{Algoritmi genetici}
Mentre le reti neurali guardano al cervello umano per produrre un comportamento intelligente, gli algoritmi genetici guardano alla teoria dell’evoluzionismo. Sono algoritmi per la ricerca nello spazio degli stati, nei quali si considera uno stato come un individuo, all’interno di una popolazione di individui che viene fatta evolvere secondo le leggi dell’evoluzionismo in modo da ottenere stati che siano buone soluzioni del problema.

      Per poter applicare un algoritmo genetico, è necessario rappresentare lo stato come una sequenza di simboli (nel caso più frequente una sequenza di bit), che rappresenta il patrimonio genetico (o genotipo) di un individuo e lo caratterizza completamente. È poi necessaria una funzione di fitness che, data una sequenza di simboli, dica quanto è “fit” l’individuo ovvero quanto è adatto a sopravvivere nel suo ambiente. Nel caso di un algoritmo genetico, la funzione di fitness rappresenta la vicinanza dello stato ad una soluzione o la sua bontà come soluzione. 

      Un algoritmo genetico parte da una popolazione iniziale composta da individui generati a caso. Esegue poi un ciclo che termina quando la fitness del miglior individuo supera una certa soglia, ovvero quando la popolazione contiene una soluzione sufficientemente buona. Ad ogni passo del ciclo si genera una nuova popolazione utilizzando gli operatori di selezione, incrocio e mutazione. In pratica, ogni iterazione del ciclo corrisponde a una generazione. La nuova popolazione è generata selezionando coppie di individui a caso ma con una probabilità che dipende dalla loro fitness: individui con fitness più alta hanno maggiore possibilità di essere selezionati. Poi si applica l’operatore d’incrocio che, data la coppia di individui, ne produce un’altra, ottenuta “mescolando” il patrimonio genetico in diversi modi: nel caso di genotipi rappresentati come sequenze di bit di lunghezza fissa $n$, l’operatore di incrocio più semplice sceglie un numero intero a caso $i$ minore di $n$ e copia nel primo discendente i primi $i$ bit del primo genitore e gli ultimi $n-i$ bit del secondo genitore,  mentre nel secondo discendente copia l’inverso. I discendenti così ottenuti sono poi soggetti a mutazione, nella quale si effettuano a caso piccoli cambiamenti del genotipo. Il processo di selezione, incrocio e mutazione viene ripetuto fino a che non si ottiene una nuova popolazione di dimensione prefissata. La fitness di tutti gli individui della nuova popolazione è poi calcolata.

      Esistono molte varianti di questo tipo di algoritmo. Ad esempio in alcune una parte della vecchia popolazione viene trasferita direttamente in quella nuova, utilizzando la selezione.

      Gli algoritmi genetici possono essere utilizzati anche per svolgere compiti di apprendimento automatico. In questo caso occorre rappresentare le descrizioni del concetto da imparare come sequenza di simboli: la fitness sarà data dall’accuratezza con la quale una descrizione del concetto classifica gli esempi del training set.

      Gli algoritmi genetici hanno avuto numerose applicazioni in biologia, ingegneria e nelle scienze fisiche e sociali. Una delle più interessanti consiste nella programmazione automatica, ovvero nella generazione automatica di programmi per calcolatore per risolvere un certo problema. In questo caso il genotipo degli individui è costituito da alberi che rappresentano un singolo programma in un dato linguaggio di programmazione. L’operatore di incrocio consiste nel sostituire un sottoalbero di un genitore con un sottoalbero dell’altro genitore. La funzione di fitness è calcolata eseguendo il programma su un insieme di dati di ingresso.

\subsection{Intelligenza degli sciami}
Gli algoritmi basati sull’intelligenza degli sciami (swarm intelligence) cercano di riprodurre il comportamento degli insetti che vivono in colonie, come formiche, api, termiti e vespe. Questi insetti sono infatti individualmente semplici ma danno luogo a comportamenti collettivi molto complessi, come ad esempio la costruzione di un formicaio e l’approvvigionamento di cibo. Si dice che il comportamento intelligente “emerga autonomamente” dal comportamento dei singoli insetti, non essendoci un supervisore o coordinatore a dirigere gli individui.

      Gli algoritmi più famosi di questa famiglia sono quelli che riproducono colonie di formiche nella ricerca del cibo. Le formiche sono in grado di trovare la strada più breve dal formicaio al cibo. Nel loro movimento ciascuna lascia sul terreno una sostanza odorosa chiamata feromone, che svanisce lentamente nel tempo; gli spostamenti puntano verso le zone dove il livello di feromone è più alto. 

      Le formiche partono inizialmente dal formicaio seguendo direzioni casuali, in quanto non c’è feromone sul terreno. La prima formica che trova il cibo e lo riporta al formicaio è quella che ha seguito il percorso più breve di andata e ritorno. Le altre che partono dal formicaio tendono quindi a seguire la strada percorsa dalla prima formica, essendo quella più odorosa. Il livello di feromone sul percorso più breve tende così a rinforzarsi, producendo quella che si chiama retroazione positiva. Dopo una prima fase in cui le formiche vagano casualmente sul terreno, si arriva ad una fase in cui tutte le formiche seguono il cammino più breve.

      Algoritmi basati sulla riproduzione di sciami sono stati efficacemente utilizzati in numerosi ambiti: ad esempio, per il problema del commesso viaggiatore, per l’indirizzamento dei pacchetti di dati nelle reti informatiche, per l’indirizzamento di veicoli nella rete stradale e per l’assegnazione di lavorazioni su macchinari.

\section{Dibattito sull'Intelligenza Artificiale}
L’IA ha scatenato un grande dibattito nella comunità scientifica e filosofica riguardo cosa significhi essere intelligenti, se le macchine potranno mai esserlo, se avranno mai una mente, se è eticamente corretto costruire macchine intelligenti…

      Questi temi sono in realtà discussi da secoli nella comunità filosofica ma recentemente hanno ricevuto molta attenzione a causa dei progressi dell’IA. Anche la letteratura e il cinema vi si sono molto interessati, come è testimoniato dalla grande quantità di libri e film in cui compaiono robot – umanoidi o meno – e macchine intelligenti. Terminator 2 - Il giorno del giudizio di James Cameron (1991) si apre con la scritta “Los Angeles, 2029 A.D.” mentre sullo sfondo appaiono le immagini di una guerra tra umani e robot, scatenata dal cervello elettronico che comanda tutti i dispositivi di difesa della Terra, Skynet, diventato autocosciente e sfuggito al controllo dei suoi creatori. Skynet è basato su un potente processore neurale e ha impiegato 25 giorni a diventare autocosciente, attraverso un velocissimo processo di apprendimento.

      I filosofi distinguono due ipotesi: l’IA debole e l’IA forte. L’IA debole afferma che le macchine sono (o saranno) in grado di comportarsi come se fossero intelligenti, ovvero di risolvere tutti i problemi che l’intelligenza umana sa risolvere. L’IA forte invece afferma che le macchine sono (o saranno) in grado di pensare, ovvero di avere una intelligenza indistinguibile dalla mente umana.

      Varie critiche sono state mosse a entrambe le ipotesi. 
      Se consideriamo l’IA debole, molti filosofi hanno affermato: “le macchine non saranno mai in grado di svolgere il compito X” dove X è stato di volta in volta “battere un umano a scacchi”, “scrivere musica”, “riconoscere il parlato”, “fare scoperte scientifiche”, “superare il test di Turing” e altri. In molti casi queste affermazioni sono state smentite dai fatti: una macchina ha battuto il campione del mondo di scacchi, brani musicali generati dal computer sono stati ritenuti indistinguibili da brani composti da un essere umano, la lingua parlata viene compresa dai computer e piccole ma significative scoperte sono state fatte dai computer in matematica, astronomia, chimica, mineralogia, biologia e informatica. È difficile quindi affermare ora con certezza che alcuni compiti non potranno mai essere svolti da una macchina.

      Computer che superano il test di Turing invece non sono ancora stati costruiti: ogni anno dal 1991 viene tenuta una competizione chiamata Loebner Prize in cui i partecipanti sono macchine che vengono sottoposte al test. Il costruttore della prima macchina che riuscirà a superarlo riceverà 100.000 dollari. Ciascun giudice, dopo la conversazione, assegna all’interlocutore un punteggio da 1 a 10 dove 1 significa computer e 10 essere umano. I punteggi medi attuali si stanno avvicinando sempre più a 5.

      In merito all’IA forte, invce, come facciamo a sapere se una macchina ha un’intelligenza indistinguibile da un essere umano? Abbiamo visto che la macchine svolgono un numero sempre maggiore di compiti in passato esclusivi degli uomini, si può quindi ipotizzare che, un giorno potranno risultare indistinguibili da un essere umano, a meno che non vengano nel frattempo scoperte impreviste limitazioni tecniche. Ma per essere effettivamente indistinguibili dovranno, secondo molti filosofi, divenire autocoscienti, ovvero consapevoli dei propri stati mentali e delle proprie azioni, come gli umani, ed essere dotate di libero arbitrio.

      Alcuni filosofi escludono questa possibilità dal momento che i comportamenti delle macchine sono regolati da leggi fisiche deterministiche che non lasciano spazio a scelte. Questa obiezione non è valida se si crede nel materialismo, ovvero nella teoria secondo la quale non esiste un’anima immateriale separata dal corpo, ma solo oggetti materiali, e di conseguenza ogni stato mentale non è altro che uno stato del cervello regolato da leggi fisiche (di natura diversa da quelle che regolano i computer ma pur sempre leggi fisiche): ogni neurone risponde agli stimoli di ingresso in una maniera preordinata secondo precise leggi elettrochimiche. Nonostante questo, noi siamo capaci di libero arbitrio, di provare emozioni e di comportarci irrazionalmente. Ciò significa che dobbiamo rivedere la nostra nozione ingenua di libero arbitrio e pensare che sistemi molto complessi come il cervello possano manifestare proprietà olistiche, ovvero che si applicano al sistema nell’insieme ma non alle singole parti che lo compongono. 

      Se si crede nel materialismo, quindi, non si può escludere che una macchina, un giorno, diventi autocosciente. Questo è vero a maggior ragione sulla base dei recenti sviluppi delle reti neurali artificiali, che mirano proprio a riprodurre il funzionamento fisico del cervello umano. 

      Se non si crede nel materialismo, si crede nel dualismo, inizialmente proposto da Cartesio (1596-1650), che prevede un’anima separata dal corpo materiale. Questa è anche la teoria di molte religioni, secondo le quali l’anima è una proprietà del solo uomo, donata da Dio e immortale, ed è essa che determina la coscienza e il libero arbitrio. In questo caso chiaramente non si ammette che una macchina possa avere queste stesse proprietà.

      Secondo i filosofi e gli scienziati che non hanno convinzioni religiose, il dualismo è però attaccabile sotto vari aspetti \cite{but}. Prima di tutto, se la mente è separata dall’uomo, come fa a comandarne le azioni, ovvero come fa a interagire con i tessuti dell’uomo per provocare quelle reazioni chimiche che risultano in un’azione fisica. Per ammettere il dualismo, occorrerebbe quindi negare le leggi fondamentali della fisica.

      In secondo luogo, se la mente è indipendente dal cervello, perché noi abbiamo un cervello?

      Terzo, se i pensieri e le emozioni ci derivano dall’esterno, perché un cervello stimolato per mezzo di elettrodi e droghe reagisce generando pensieri?

      Quarto e ultimo, perché la rimozione chirurgica di parti del cervello in pazienti affetti da malattie cerebrali porta a una alterazione del loro comportamento? 

       Supponendo quindi di credere nel materialismo e che in futuro non si scoprano difficoltà tecniche che limitino le capacità delle macchine, ci si può chiedere quando una macchina diventerà autocosciente. Secondo il materialismo, una rete neurale complessa quanto il cervello umano dovrebbe presentarne le stesse proprietà, tra cui l’autocoscienza e il libero arbitrio. Giorgio Buttazzo in \cite{but} ha calcolato la data in cui un personal computer sarà in grado di simulare una rete neurale complessa quanto il cervello umano. Quest’ultimo possiede ha circa mille miliardi di neuroni ($10^{12}$), ogni neurone ha circa mille ($10^3$) connessioni con gli altri neuroni (sinapsi), per un totale di $10^{15}$ sinapsi. Abbiamo visto che ad ogni ingresso di un neurone è associato un peso che ha valore reale. Per rappresentare un numero reale in un calcolatore occorrono 4 bytes di memoria. Di conseguenza per simulare $10^{15}$ sinapsi occorre una memoria di almeno $4\cdot 10^{15}$ bytes ovvero 4 milioni di Gigabytes. Stimando in un milione di Gigabytes la memoria necessaria a memorizzare lo stato dei neuroni più le altre variabili ausiliarie necessarie per la simulazione, si ottiene un totale di 5 milioni di Gigabytes. Quando sarà disponibile una tale memoria per personal computer?

      Fino ad oggi le capacità dei calcolatori sono state descritte dalla legge enunciata nel 1973 da Gordon Moore, uno dei fondatori dell’Intel, secondo cui il numero dei transistor sarebbe raddoppiato ogni 18 mesi fino al raggiungimento dei limiti fisici. Questo equivale a una moltiplicazione per un fattore 10 ogni 4 anni. Dato che la capacità delle memorie è una funzione lineare del numero dei transistor, per la capacità $C$ in bytes della memoria RAM di un PC si può scrivere la seguente legge
   $$C=10^{(\mathit{Anno}-1966)}/4$$
      Questa legge è stata valida dall’introduzione dei personal computer nel 1980 fino ad oggi. Supponendo che rimanga valida anche nel futuro, l’anno in cui si otterrà un certo valore di capacità $C$ è dato da
      $$\mathit{Anno}=1966+ 4 \log_{10} C$$
      Per $C$ uguale a $5\cdot 10^6$ Gigabytes si ottiene Anno = 2029, esattamente l’anno in cui, in Terminator 2, Skynet prende coscienza! 

      Ci dobbiamo ora chiedere se valga la pena di costruire macchine del genere. Contro l’IA vi sono varie critiche, la più terrificante delle quali fa riferimento a scenari appunto quali quello di Terminator 2, in cui le macchine prendono il controllo e cercano di annullare la razza umana. Dire oggi se questi scenari si potranno realizzare è molto difficile: data la limitatezza delle capacità dei sistemi attuali, prevedere se sarà possibile realizzare sistemi autocoscienti (al di la di pronostici azzardati come quello sopra) è molto difficile. Supponendo poi che ci si possa riuscire, cosa ci fa pensare che questi sistemi debbano produrre la stessa aggressività mostrata dall’uomo nel corso della sua storia? Inoltre, la scienza finora ha proceduto per passi incrementali, cosa ci fa pensare che nel futuro si verificheranno delle discontinuità del progresso che porteranno l’uomo a non accorgersi del pericolo e a perdere il controllo? 

Nick Bostrom nel suo libro \cite{bostrom2018superintelligenza} considera proprio il caso in cui le macchine raggiungano l'intelligenza umana ed afferma che le macchine con tale intelligenza saranno l'ultima invenzione che l'essere umano dovrà
mai fare. Infatti, una volta che una macchina abbia raggiunto il livello umano e l'autocoscienza, potrà rapidamente migliorare se stessa raggiungendo così una superintelligenza. Il libro illustra  vari scenari e conclude the difficilmente
sarà possibile controllare una tale macchina.  Una soluzione al problema del controllo è proposta da Stuart Russell nel suo libro \cite{russell2019human}: Russell suggerisce di creare macchine il cui scopo sia quello di soddisfare i bisogni
umani ma di lasciare alla macchina il compito di capire quali siano questi bisogni, includendo in particolare incertezza su questi obiettivi. In questo modo le macchine dovranno osservare il comportamento umano per capire i bisogni
umani e raggiungere in questo modo il loro scopo. Dovendo osservare gli esseri umani, le macchine non cercheranno di eliminare la razza umana.

      A favore dell’IA si può annoverare la possibilità di aumentare enormemente le capacità dell’uomo di conoscere il mondo circostante, di costruire beni e offrire servizi. Alla luce dei problemi di sostenibilità che il nostro stile di vita presenta, l'IA
potrebbe fornirci un aiuto fondamentale per ottimizzare le risorse di cui disponiamo, avvicinandoci a una società maggiormente sostenibile. 
      
      Inoltre le macchine possono rispondere ad un nostro desiderio di immortalità: possono rappresentare l’opportunità di trasferire la mente umana su un supporto più duraturo, ad esempio utilizzando le reti neurali artificiali. In questo modo verremmo liberati dalle limitazioni imposte dalla nostra natura biologica. Si tratterebbe della nascita di una nuova umanità, forse il passo successivo della nostra evoluzione biologica. C’è addirittura un movimento, chiamato transumanismo, che spera in questo futuro. Ovviamente si tratta di un futuro che la maggior parte dei teorici della morale aborrisce, considerando la conservazione della vita umana come un bene supremo. Un tale trasferimento della mente da un substrato biologico ad un substrato tecnologico potrebbe essere però eseguito come ultima risorsa prima della morte di un individuo.

Anche se questi sono  problemi che è  presto porsi perché siamo ancora lontani dal costruire macchine con intelligenza paragonabile a quella umana, il dibattito sull'etica dell'IA è già cominciato in quanto l'IA sta già avendo
un'influenza significativa sulla vita umana. A questo scopo, la Comunità Europea ha costituito un gruppo di esperti che ha prodotto un documento contenente linee guida per una IA degna di fiducia o “trustworthy”. In questo documento
il gruppo sottolinea l'importanza di un approccio all'IA che sia centrato sull'uomo e propone un elenco di requisiti affinché un sistema di IA possa essere considerato degno di fiducia.

\bibliographystyle{alphaurl}
\bibliography{ia}

\begin{thebibliography}{CLMM97}

\bibitem[CAD04]{car}
L.~Carlucci~Aiello and M.~Dapor.
\newblock Intelligenza artificiale: i primi 50 anni.
\newblock {\em Mondo Digitale}, 10, giugno 2004.
\newblock URL:
  \url{http://archivio-mondodigitale.aicanet.net/Rivista/04_numero_tre/Aiello,%20Dapor%20p.%203-20.pdf}.

\bibitem[CLMM97]{con}
L.~Console, E.~Lamma, P.~Mello, and M.~Milano.
\newblock {\em Programmazione logica e Prolog}.
\newblock UTET, 1997.

\bibitem[Gor03]{gor}
M.~Gori.
\newblock Introduzione alle reti neurali artificiali.
\newblock {\em Mondo Digitale}, 8, dicembre 2003.
\newblock URL:
  \url{http://archivio-mondodigitale.aicanet.net/Rivista/04_numero_uno/Gori_p.4-20.pdf}.

\bibitem[Odi04]{odi}
P.~Odifreddi.
\newblock {\em Il diavolo in cattedra. La logica da Aristotele a Gödel}.
\newblock Collana Einaudi Tascabili, Saggi. Einaudi, 2004.

\end{thebibliography}


\begin{thebibliography}{Hau85}

\bibitem[Bos18]{bostrom2018superintelligenza}
N.~Bostrom.
\newblock {\em Superintelligenza: Tendenze, pericoli, strategie}.
\newblock Bollati Boringhieri Saggi. Bollati Boringhieri, 2018.

\bibitem[But02]{but}
G.~Buttazzo.
\newblock Coscienza artificiale: Missione impossibile?
\newblock {\em Mondo Digitale}, 1:16--25, marzo 2002.
\newblock URL:
  \url{http://archivio-mondodigitale.aicanet.net/Rivista/02_numero_uno/Buttazzo.pdf}.

\bibitem[Cyb89]{DBLP:journals/mcss/Cybenko89}
George Cybenko.
\newblock Approximation by superpositions of a sigmoidal function.
\newblock {\em Math. Control. Signals Syst.}, 2(4):303--314, 1989.
\newblock \href {https://doi.org/10.1007/BF02551274}
  {\path{doi:10.1007/BF02551274}}.

\bibitem[Hau85]{10.5555/4694}
John Haugeland.
\newblock {\em Artificial Intelligence: The Very Idea}.
\newblock Massachusetts Institute of Technology, USA, 1985.

\bibitem[Mel02]{mel}
P.~Mello.
\newblock Intelligenza artificiale.
\newblock In {\em Dizionario interdisciplinare di Scienza e Fede}. Urbaniana
  University Press, 2002.

\bibitem[Mit97]{mit}
T.~M. Mitchell.
\newblock {\em Machine Learning}.
\newblock McGraw-Hill, 1997.

\bibitem[MP43]{mccpit}
W.~S. McCulloch and W.~Pitts.
\newblock A logical calculus of the ideas immanent in nervous activity.
\newblock {\em Bullettin of Mathematical Biophysics}, 5(4):115--137, 1943.

\bibitem[RN05]{rusnor}
S.~Russell and P.~Norvig.
\newblock {\em Intelligenza artificiale. Un approccio moderno, 2a edizione}.
\newblock Pearson Education Italia, 2005.

\bibitem[Rus19]{russell2019human}
S.~Russell.
\newblock {\em Human Compatible: AI and the Problem of Control}.
\newblock Penguin Books Limited, 2019.

\bibitem[SB12]{deep_learn}
Hannes Schulz and Sven Behnke.
\newblock Deep learning.
\newblock {\em KI - K{\"u}nstliche Intelligenz}, 26(4):357--363, 2012.
\newblock \href {https://doi.org/10.1007/s13218-012-0198-z}
  {\path{doi:10.1007/s13218-012-0198-z}}.

\bibitem[Sim84]{sim}
H.~A. Simon.
\newblock Why should machines learn.
\newblock In R.~S. Michalski, J.~G. Carbonell, and T.~M. Mitchell, editors,
  {\em Machine Learning - An Artificial Intelligence Approach}, page 25—37.
  Springer-Verlag, 1984.

\bibitem[Tur50]{tur}
A.~Turing.
\newblock Computing machinery and intelligence.
\newblock {\em Mind}, 59:433--460, 1950.

\end{thebibliography}

\bibliographystylealtre{alphaurl}
\bibliographyaltre{ia}

\section{Collegamenti}
\begin{itemize}
\item
Associazione Italiana per l'IA:  \url{http://www.aixia.it/}
\item
Associazione Europea per l'IA: \url{http://www.eurai.org/}
\end{itemize}

\end{document}